\pdfoutput=1

\documentclass[11pt]{article}

\usepackage[]{acl}

\usepackage{times}
\usepackage{latexsym}

\usepackage{multicol}
\usepackage{multirow}
\usepackage{graphicx}

\usepackage[T1]{fontenc}

\usepackage[utf8]{inputenc}

\usepackage{microtype}

\title{Scientific Fact-Checking: A Survey of Resources and Approaches}

\author{Juraj Vladika \and Florian Matthes \\
  Department of Computer Science \\ Technical University of Munich \\ Garching, Germany \\
  \texttt{ \{juraj.vladika, matthes\}@tum.de} \\}

\begin{document}

\maketitle
\begin{abstract}
The task of fact-checking deals with assessing the veracity of factual claims based on credible evidence and background knowledge. In particular, scientific fact-checking is the variation of the task concerned with verifying claims rooted in scientific knowledge. 
This task has received significant attention due to the growing importance of scientific and health discussions on online platforms. Automated scientific fact-checking methods based on NLP can help combat the spread of misinformation, assist researchers in knowledge discovery, and help individuals understand new scientific breakthroughs. In this paper, we present a comprehensive survey of existing research in this emerging field and its related tasks. We provide a task description, discuss the construction process of existing datasets, and analyze proposed models and approaches. Based on our findings, we identify intriguing challenges and outline potential future directions to advance the field.
\end{abstract}

\section{Introduction}
In today's digital age, vast amounts of data are generated and new scientific breakthroughs achieved at a rapid pace. With millions of scientific articles being published annually, it has become increasingly challenging for researchers and the general public to stay informed about the latest developments and discoveries across various fields. On top of that, an especially challenging task for researchers is finding appropriate evidence for scientific claims and research hypotheses they are currently investigating. Exploring large academic databases and thoroughly examining scientific publications in them in order to verify specific facts is a time-consuming process. Automating the process of fact-checking scientific claims using methods based on Natural Language Processing (NLP) for knowledge exploration and evidence mining can greatly aid researchers in these efforts.

One way how the Internet has benefited society is by making scientific knowledge easily accessible, transferable, and searchable in a matter of seconds. Inevitably, this has introduced new risks and challenges -- it has become difficult to discern reliable sources from dubious content. Many scientific claims found in online articles, social media posts, or news reports are not always trustworthy and backed by reliable evidence. Furthermore, not only are humans prone to creating inaccurate information -- modern generative language models can also produce misleading text that sounds convincing. All of these factors, combined with the quick pace at which content is proliferated online, contribute to the spread of misinformation, which has negative societal consequences \citep{West2021MisinformationIA}.

Fact-checking is the task of assessing the veracity of factual claims appearing in written or spoken sources. It is traditionally performed manually by experts in journalism and dedicated applied fields. Automated fact-checking appeared as an approach where methods of Natural Language Processing (NLP) and Machine Learning (ML) are used to assist experts in making these decisions or completely automating the whole process \citep{Nakov2021AutomatedFF}. Fact-checking becomes especially relevant during major political events like elections or referendums because of a sharp increase in deceptive and propagandist content. Most recently, the COVID-19 pandemic has brought the scientific discourse and the misinformation that comes with it into the spotlight. Medical misinformation is especially dangerous because it has influenced people to try unproven cures and treatments and make harmful health-related decisions. \citep{doi:10.1098/rsos.201199, doi:10.1177/0956797620939054}.

We define \textit{scientific fact-checking} as a subset of the fact-checking task concerned with verifying the veracity of claims related to scientific knowledge. While the primary role of general fact-checking is to help detect misinformation and curb its spread, scientific fact-checking additionally aids scientists in testing their hypotheses and helps wider audiences contextualize new scientific findings. The most popular scientific domain in scientific NLP research is the biomedical domain \citep{rajpurkar2022ai}, but insights learned from it can be generalized to other scientific domains. Scientific fact-checking can be performed both over the highly structured and complex language of science found in research publications and over the more easily understandable language found in news articles and online postings meant for lay audiences. Many scientists have decried the misinterpretation of their work when presented in the news press \citep{10.1371/journal.pmed.1001308}, which makes scientific fact-checking even more relevant in bridging the gap between these two registers by performing an evidence-based assessment of scientific discoveries.


Considering the constantly increasing amount of misinformation in the digital era and the expanding number of scientific publications, the interest in developing automated fact-checking solutions and efficient resources for it is on the rise. We present this survey to systematize the existing work in this area. To the best of our knowledge, this is the first survey on fact-checking with a specific focus on the scientific domain. Our three main contributions are:
\begin{enumerate}
  \item We describe existing datasets for scientific fact-checking, including their construction process and main characteristics.
  \item We analyze the developed approaches and models for solving the task of scientific fact-checking, focusing on their components and design choices.
  \item We outline general findings, identify challenges, and highlight promising future directions for this emergent task.
\end{enumerate}

\section{Task Definition}
\subsection{General Fact-checking}
In general, \textit{fact-checking} can be defined as the task of assessing whether a factual claim is valid based on evidence. It is a time-consuming task that is still usually performed manually by journalists. Automated approaches based on NLP have emerged to help assist humans in parts of the fact-checking process. Popular datasets used for benchmarking this task in NLP contain rewritten Wikipedia sentences as claims and annotated articles as evidence \citep{thorne-etal-2018-fact, jiang-etal-2020-hover}. For real-world settings, datasets were constructed by collecting claims and expert-written verdicts from dedicated fact-checking websites, such as PolitiFact \citep{vlachos-riedel-2014-fact}, Snopes \citep{hanselowski-etal-2019-richly}, or MultiFC \citep{Augenstein2019MultiFCAR} which draws from 26 fact-checking portals. This type of datasets usually contains claims currently trending in society, related to topics from world news, politics, media, or online rumors and hoaxes.

\subsection{Scientific Fact-checking}
We define \textit{scientific fact-checking} as a variation of the fact-checking task that deals with assessing claims rooted in scientific knowledge. The dominant purpose of general fact-checking is to combat the spread of misinformation, while scientific fact-checking has the additional motive of helping scientists verify their research hypotheses, discover evidence, and facilitate scientific work. Scientific fact-checking comes with specific challenges not always present in general fact-checking, such as:

\begin{itemize}
  \item \textbf{Claims:} Facts to be checked can be research hypotheses that scientists want to verify, claims made by everyday social media users, or queries posed to search engines dealing with scientific concepts (e.g., health-related concerns). 
  \item \textbf{Evidence:} Scientific knowledge is constantly evolving when new research is conducted, which can make previous evidence obsolete and invalid. Moreover, different studies can come to diverging conclusions which complicates the final assessment of a claim. In clinical settings, this obstacle is facilitated by systematic reviews, which provide levels of evidence and strength of recommendations for any decision. 
  \item \textbf{Domain:} The scientific language used in research publications is highly complex and contains domain-specific terminology, which presents a challenge for a general-purpose language model. This requires adapting the NLP systems to the scientific domain. On top of that, scientific text often contains relations between concepts spanning multiple sentences, which makes representation of the full context and long-text modeling an essential aspect.
  \item \textbf{Structure:} The highly structured nature of scientific knowledge makes it convenient to model it with structured representations like knowledge graphs, which can aid the fact-checking process. On the other hand, scientific publications commonly include different visualization techniques like tables, charts, and figures, all of which introduce additional multimodal challenges to verification.
\end{itemize}

\begin{table*}[t]
\centering
\small
\begin{tabular}{l|clll}
\hline
\textbf{Dataset}  & \textbf{\# Claims} & \textbf{Claim Origin   }            & \textbf{Evidence Source }                       & \textbf{Domain} \\ \hline \\
\textsc{SciFact}   \citep{wadden-etal-2020-fact}       & 1,409     & Researchers          & Research papers               & Biomedical             \\[3ex]
\textsc{PubHealth}  \citep{kotonya-toni-2020-explainable}    & 11,832    & Fact-checkers & Fact-checking sites          & Public health             \\[3ex] 
 \textsc{Climate-FEVER} \citep{Diggelmann2020CLIMATEFEVERAD} & 1,535     & News articles              & Wikipedia articles                     & Climate change             \\[3ex] 
\textsc{HealthVer}  \citep{Sarrouti2021Healthver}    & 1,855    & Search queries      & Research papers               & Health            \\[3ex]
\textsc{COVID-Fact} \citep{saakyan-etal-2021-covid}      & 4,086     & Reddit posts               & Research, news & COVID-19             \\[3ex]
\textsc{CoVERT}    \citep{mohr-whrl-klinger:2022:LREC}     & 300       & Twitter posts              & Research, news & Biomedical \\[2ex]
\hline
\end{tabular}

\caption{\label{tab:datasets}Datasets for the task of scientific fact-checking and claim verification}
\end{table*}

These characteristics and other challenges with scientific fact-checking will be discussed in more detail in the following sections, especially in the Discussion section.

\section{Related Tasks}
In this section, we present tasks related to scientific fact-checking. We group them into three categories: (1) tasks related to misinformation detection; (2) retrieval of claims, arguments, and evidence from text; and (3) NLP tasks in the scientific domain.

\subsection{Misinformation Detection}
Since the principal function of fact-checking is to curb the spread of misinformation, it naturally belongs to a group of NLP tasks concerned with misinformation detection. Related tasks in this domain include fake news detection \citep{10.1145/3395046}, propaganda detection \citep{10.5555/3491440.3492112}, rumor detection \citep{Bian2020RumorDO}, or stance detection \citep{hardalov-etal-2022-survey}. While most of these tasks deal with misinformation related to politics and society, recently, there has been an increase in scientific and health-related misinformation detection, especially pertaining to content related to the COVID-19 pandemic \citep{Shahi_Nandini_2020, hossain-etal-2020-covidlies, antypas-etal-2021-covid}.


\subsection{Claim Detection and Evidence Mining}

%

A crucial prerequisite for automated fact-checking is devising methods that detect claims in the open domain. To achieve this, \citet{10.1016/j.ipm.2019.03.001} used a rule-based system to identify health claims in news headlines, while \citet{wuhrl-klinger-2021-claim} develop a BERT-based model to detect biomedical claims in social media posts. 
After the claims are detected, an important next step is determining whether a claim is check-worthy since all claims are deemed relevant or interesting enough to be fact-checked. Check-worthiness for scientific claims was studied in the shared task CLEF-CheckThat! \citep{10.1007/978-3-031-13643-6_29} and by \citet{zuo-journal-2022}, where annotators helped construct a dataset of health-related claims from news articles.

Automatic gathering of evidence for scientific claims constitutes another line of research. There is work in this area focusing on humanities and social sciences \citep{tubiblio127669}, although the majority of work we found is once again in life sciences. Numerous tools have been developed for searching PubMed, the largest database of biomedical publications \citep{zhiyong2011}, such as PubTator \citep{10.1093/nar/gkz389}, Textpresso \citep{textpresso}, LitSense \citep{litsense}, and EvidenceMiner \citep{wang-etal-2020-evidenceminer}. These methods usually look at the posed query (claim) and detect named entities, keywords, or metadata patterns to retrieve relevant results from the database. The end goal of this process is to help scientists gather evidence for their research, while in fact-checking, evidence retrieval is just one component of the whole process.

\subsection{Scientific NLP Tasks}
Scientific fact-checking belongs to a group of NLP tasks dealing with scientific text understanding. These tasks share a common challenge: working with highly complex scientific language and specific terminology. This has become even more apparent with the underwhelming performance of large language models, pre-trained on vast amounts of news data and web content, on NLP tasks in the scientific domain. Domain adaption is an essential cornerstone of modern NLP models working with specialized domains. 

The task of Natural Language Inference (NLI), commonly equated with Recognizing Textual Entailment (RTE), is the task of inferring whether a premise entails or contradicts a given hypothesis. This task is a crucial component of automated fact-checking since predicting the final veracity of the claim is modeled entailment recognition between a claim and found evidence. For the scientific domain, datasets like MedNLI, which features medical claims rooted in the medical history of patients \citep{romanov2018lessons}; SciNLI, which has claims from the domain of computational linguistics \citep{sadat2022scinli}; and NLI4CT, with claims and evidence that originate from clinical trials reports of breast cancer patients \citep{vladika2023sebis}.
 
Another knowledge-intensive NLP task related to fact-checking is question answering. In particular, open-domain question answering aims to find answers to given questions in unstructured textual corpora \citep{karpukhin-etal-2020-dense}, reminiscent of the process of finding relevant evidence for given claims in fact-checking. Popular datasets for biomedical QA are BioASQ \citep{tsatsaronis2015overview} and PubMedQA \citep{jin2019pubmedqa}. Another important benchmark is BLURB (Biomedical Language Understanding and Reasoning Benchmark), introduced by \citet{Gu_2022} to measure the performance of models in six different natural language understanding tasks over biomedical text. Finally, automated evidence synthesis is a task that aims to automate the process of creating systematic reviews for clinical trials \citep{brassey2021developing}.

\subsection{Related Surveys}
There are already existing surveys that cover general automated fact-checking \citep{thorne-vlachos-2018-automated, zeng2021, guo-etal-2022-survey} by formalizing the task, outlining the most important datasets and proposed solutions, and discussing challenges. The survey by \citet{kotonya-toni-2020-explainable-survey} focuses on explainability methods in existing fact-checking approaches and present the most important explainability aspects these systems should satisfy. The survey by \citet{bekoulis2021} focuses on approaches for tackling FEVER, the most popular dataset for fact verification \citep{thorne-etal-2018-fact}.

\section{Datasets}
In this section, we outline the existing datasets for scientific fact-checking that we found in the literature. The discovery process started with querying the well-known databases ACL Anthology,\footnote{\url{https://aclanthology.org/}} IEEE Explore,\footnote{\url{https://ieeexplore.ieee.org/}} and ACM Digital Library\footnote{\url{https://dl.acm.org/}}  with the search string \textit{("scientific" OR "biomedical") AND ("fact checking" OR "fact verification" OR "claim verification")}. Retrieved articles were collected and the list was further expanded with any cited or citing paper from the initial batch of articles, according to Semantic Scholar.\footnote{\url{https://www.semanticscholar.org//}} In order for a dataset to be considered a fact-checking dataset, we stipulate it needs to provide claims, evidence (either documents or sentences), and final veracity labels. Such a dataset enables both the task of evidence retrieval and verdict prediction. This is important because the end goal of many automated fact-checking systems is to emulate the work of experts, where both seeking the evidence and making conclusions based on them constitute the process. This requirement narrowed the final list to the datasets summarized in Table \ref{tab:datasets}. In the remainder of the section, we will describe the process and challenges related to constructing datasets.


\subsection{Claim Creation}
The starting point in the dataset construction process is collecting the claims that will later be fact-cheked. Claims in fact-checking are usually divided into synthetic, referring to claims written by annotators (e.g., by modifying sentences from Wikipedia), and natural, which are those claims crawled from real-world sources like fact-checking sites or social media posts. The first type of claims end up being fluent, atomic, and decontextualized, which is very appropriate for processing by NLP models \citep{wright-etal-2022-generating}. Other authors focus on more organic and noisy claims found in online posts since such claims are usually relevant and interesting to be fact-checked automatically \citep{mohr-whrl-klinger:2022:LREC}.

One common approach is to take original sentences from an appropriate source and have annotators reformulate them to a cleaner form. The dataset \textsc{SciFact} features biomedical claims that originate from human-written citation sentences in research articles but with the final form rewritten by annotators to make them more atomic and easily processed. Similarly, claims in \textsc{HealthVer} originate from Bing snippets of the most-searched user queries related to health and COVID-19, eventually reformulated by annotators. In the same vein, \textsc{Climate-FEVER} contains sentences related to climate change extracted from online blogs and news websites, rewritten by annotators.

The remaining datasets from Table \ref{tab:datasets} relied on completely automatically retrieving claims. \textsc{PubHealth} used news titles from fact-checking articles related to public health as its claims. This assumption works in many cases where titles are indeed factual claims, but some examples in the final dataset are generic titles with no relevance for fact-checking. \textsc{COVID-Fact} scraped claims from posts of a highly moderated subreddit \textit{r/COVID19}, where users were already required to make atomic claims in their post titles. They also automatically constructed all of their negative (refuted) claims with word in-filling from masked language models, which ended with some unusable examples. Finally, \textsc{\textsc{CoVERT}  } is the only dataset in the list that features completely organic claims found in Twitter posts. They used a biomedical claim detection model \citet{wuhrl-klinger-2021-claim} to extract claims that feature a causative relation and also included mentions of any biomedical entities.


\subsection{Evidence Set Construction}
Once the claims are collected, the next step is pairing them with appropriate evidence that addresses their veracity. The evidence source are often scientific publications, featuring highly complex and structured scientific language, or more easily understandable sources like news articles and Wikipedia articles. While working with text from scientific publications is more challenging both for humans and NLP models alike, they provide more rigorous scientific evidence. On the other hand, the general-purpose text provides evidence in a more explainable and intuitive form to a wider audience.

The \textsc{SciFact} dataset pairs the claims with abstracts of those scientific publications where they originated from, adding distractor abstracts to make detecting appropriate evidence more challenging. Likewise, claims in \textsc{HealthVer} are also mapped to appropriate scientific publications found by the annotators. Datasets \textsc{COVID-Fact} and \textsc{\textsc{CoVERT}  } feature a combination of both scientific publications and news articles as their evidence source, while \textsc{PubHealth} uses solely the web articles from fact-checking websites where their claims originated from. In the same way as the original \textsc{FEVER} dataset,   \textsc{Climate-FEVER} uses Wikipedia articles as its evidence source.


\subsection{Class Labels}
Another integral component of dataset construction is labeling the claims with appropriate veracity labels. Following the tradition set by the \textsc{FEVER} dataset \cite{thorne-etal-2018-fact}, most of the datasets include three labels: \textsc{Supported}, \textsc{Refuted}, and \textsc{Not Enough Information} (NEI). The definition of the NEI label has a different meaning in different datasets. In \textsc{\textsc{SciFact} }, this label refers to those claims for which none of the candidate abstracts contain suitable evidence to make a decision. In other datasets, it refers to the case where relevant evidence itself implies or states that there is currently not enough information to make a reliable and informed conclusion about the claim's veracity. Additionally, the dataset \textsc{PubHealth} is the only one to feature a \textsc{Mixed} label, a label denoting a claim that consists of multiple factual statements with opposite veracity labels.

\section{Approaches}

\begin{figure}[htpb]
  \centering
  \includegraphics[width=0.99\linewidth]{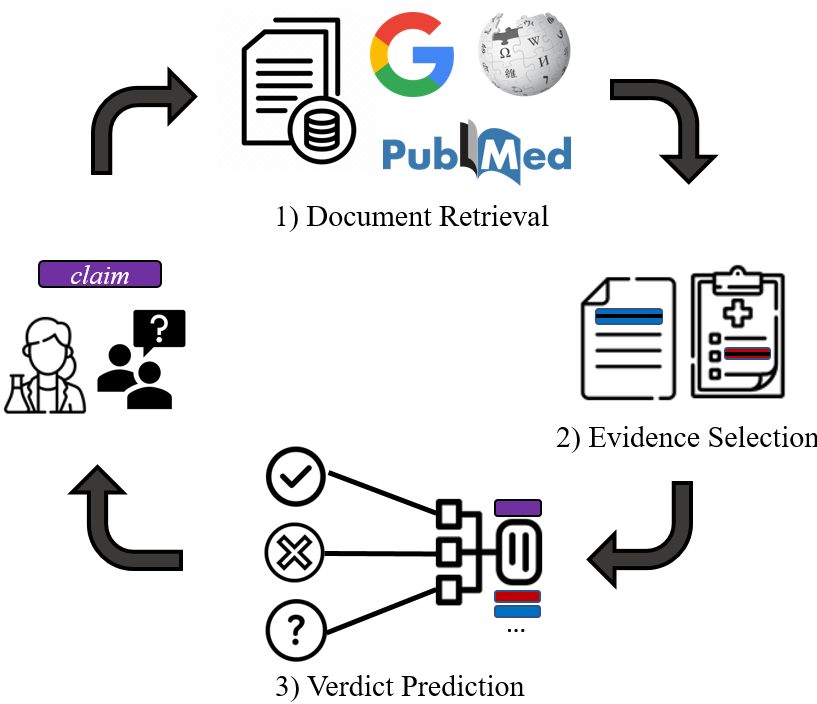}
  \caption{The standard three components of the framework for automated scientific fact-checking}
  \label{fig:dc}
\end{figure}

In this section, we describe different modeling approaches devised for the task of scientific fact-checking. The standard framework usually consists of three major components that can all be modeled as well-established NLP tasks: document retrieval, evidence (rationale) selection, and verdict prediction \citep{zeng2021}. This framework is visualized in Figure \ref{fig:dc}. 

Table \ref{tab:models} summarizes the models we found in the literature, developed for the scientific fact-checking datasets from the previous chapter, with three framework components in each of them highlighted. While the most common approach is building separate models for each element and applying them in a pipeline, the best-performing systems jointly learn the rationale selection and verdict prediction with a shared representation.
The dataset \textsc{SciFact} has the most models developed for it, partly owing to the shared task SCIVER \citep{wadden-lo-2021-overview}. For some of the datasets, we did not find dedicated models other than baselines provided in their originating papers. We analyze each part of the framework in more detail.

\begin{table*}[t]
\centering
\begin{tabular}{p{19mm} p{29mm} p{25mm} p{25mm} p{25mm}p{11mm} }
\hline
\textbf{Dataset} & \textbf{Model} &\vbox{\hbox{\strut \textbf{Document}}\hbox{\strut \textbf{Retrieval}\vspace{-1ex}}} & \vbox{\hbox{\strut \textbf{Rationale}}\hbox{\strut \textbf{Selection}}\vspace{-1ex}} & \vbox{\hbox{\strut \textbf{Verdict}}\hbox{\strut \textbf{Prediction}}\vspace{-1ex}} & \textbf{Result (F1)} \\ \hline

\multirow{11}{*}{\textsc{ SciFact} }           & VeriSci \citep{wadden-etal-2020-fact}        &  TF-IDF                      & BERT                                    & BERT   &        0.395                                     \\ \cline{2-6}
         & ParagraphJoint \citep{li2021paragraph} &   BioSentVec                  & BERT + MLP / BERT + KGAT                & BERT + MLP       & 0.609                  \\ \cline{2-6}
 & VerT5erini \citep{pradeep-etal-2021-scientific}     & BM25 + T5 re-ranker (tuned on MS MARCO)        & T5 (tuned on MS MARCO)                        & T5 (no fine-tuning)    & 0.634                       \\[4ex] \cline{2-6}
   & ARSJoint   \citep{ARSJoint}    &  BioSentVec                  & BioBERT, MLP             & BioBERT, MLP          &  0.655             \\ \cline{2-6}
  & MultiVerS    \citep{wadden-etal-2022-multivers}  & BM25 + T5 re-ranker         & Longformer (binary head) & Longformer (ternary head)     & 0.672     \\ \cline{2-6} \hline
\textsc{CoVERT}          & Zero-shot MultiVerS  \citep{wuhrl-klinger-2022-entity}  &  BM25 + T5 re-ranker         & Longformer (binary head) & Longformer (ternary head)     & 0.620     \\ \hline
\textsc{PubHealth}        & Baseline  \citep{kotonya-toni-2020-explainable}     &  provided                    & Sentence-BERT                           & SciBERT      & 0.705                         \\ \hline
 \textsc{Climate-FEVER}         & ClimateBERT    \small \citep{webersinke2021climatebert} & provided                    & provided                                & ClimateBERT     & 0.757
\\ \hline
\textsc{HealthVer}        & Baseline     \citep{Sarrouti2021Healthver}  &  provided                    & provided                                & T5-base     & 0.796  \\ \hline
\textsc{COVID-Fact}       & Baseline  \citep{saakyan-etal-2021-covid}     &  Google Search               & Sentence-BERT                           & RoBERTa (fine-tuned on GLUE)       & 0.820                       \\ \hline
\end{tabular}

\caption{\label{tab:models}Models developed for scientific fact-checking with three pipeline components and verdict prediction performance on their respective dataset}
\end{table*}

\subsection{Document Retrieval}

Given a corpus of documents that serve as the knowledge source, document retrieval is concerned with retrieving the relevant documents that might contain evidence related to the claim. It is usually solved with approaches typical for Information Retrieval. These can be separated into sparse retrieval and dense retrieval approaches. Sparse retrieval uses ranking functions such as TF-IDF and BM25, which match exact keywords in a query with an inverted index of all words from the corpus. Conversely, dense retrieval deploys dense vector representations of queries, which consider the semantic meaning of the query and can catch synonyms and related concepts \citep{karpukhin-etal-2020-dense}.

For \textsc{\textsc{SciFact}}, the document retrieval task focuses on retrieving relevant abstracts from a corpus of around 5 thousand given scientific abstracts. The baseline model VeriSci uses the simple TF-IDF metric to retrieve top \textit{k} relevant abstracts. The models VerT5erini and later MultiVerS use the approach of first retrieving top \textit{k} relevant abstracts using the BM25 metric and then adjusting the rankings using a T5 \citep{10.5555/3455716.3455856} neural pointwise re-ranker based on \citep{nogueira-etal-2020-document}, which is trained on the MS MARCO passage dataset used for machine reading comprehension \citep{nguyen2016ms}. On the other hand, ParagraphJoint and ARSJoint used the dense vector representation BioSentVec \citep{Chen_2019}, which was trained from 30 million biomedical publications and clinical notes.

Searching for evidence in a small corpus of documents (5k in \textsc{\textsc{SciFact} }) is useful for experimental settings but not realistic for real-world settings where large databases with millions of scientific publications have to potentially be queried to find appropriate evidence. When expanding document retrieval for \textsc{SciFact} to 500k documents in \citep{wadden-etal-2022-scifact} and using the same BM25 + T5 re-ranking approach, the authors noticed performance drops of at least 15 points in the final F1 score of veracity prediction. This shows the need for a more precise semantic search of evidence documents. The authors of \textsc{COVID-Fact} tackle this by using snippets of the top 10 results returned by Google Search API for a given claim. This mimics how humans would approach fact-checking, but usually, additional verification of source quality and trustworthiness is needed in such an approach.

\subsection{Evidence Selection}

Evidence selection is the task of selecting relevant rationale sentences from the previously retrieved documents to be used as evidence for claim veracity prediction in the next step. Even though this step can be modeled as a span detection task, evidence is usually modeled at a sentence level. It can then be taken as a binary classification task of predicting whether a sentence is relevant or irrelevant. Most commonly, top \textit{k} sentences are selected, similarly to the document retrieval step.

A common approach to evidence selection is to deploy models for sentence similarity and take those sentences that are the most similar to the claim being checked. The baselines for \textsc{PubHealth} and \textsc{COVID-Fact} both use the Sentence-BERT model \citep{reimers2019sentence} to retrieve the top $5$ most similar sentences. Sentence-BERT is a model based on siamese networks and provides semantically rich sentence embeddings that can easily be compared using cosine-similarity. VerT5erini uses a T5 model fine-tuned on MS MARCO (same as in the previous step) for this task. 

While using sentence similarity for evidence selection is a straightforward and intuitive approach, it can fall short because evidence sentences could be paraphrased or use rather different wording from the original claim. Consequently, \citet{wright-etal-2022-modeling} improve the performance of evidence selection on \textsc{\textsc{CoVERT}  } and \textsc{COVID-Fact} datasets by fine-tuning sentence similarity models on pairs of sentences about scientific findings from scientific articles matched with paraphrased sentences from news and social media reporting on these findings. 

In all mentioned approaches, evidence selection and verdict prediction are made with two separate models, which means that the final claim veracity predictor might not have knowledge of the full context of evidence. ParagraphJoint, ARSJoint, and MultiVerS are so-called joint models because they all use multi-task learning to jointly learn the tasks of rationale selection and verdict prediction. For this purpose, they use a shared representation of the claim and the abstract obtained by concatenating the claim with the full abstract of a candidate document and converting it to a dense representation. This alleviates the problem of missing context during final label prediction. ParagraphJoint uses BERT \citep{devlin-etal-2019-bert} as the encoder model, while ARSJoint uses the domain-specific BioBERT model \citep{lee2020biobert}, pre-trained on the text of biomedical research publications. Evidence selection is performed by passing the representation of each candidate sentence (extracted from the full abstract representation) to a multi-layer perceptron (MLP) classifier. Likewise, MultiVerS obtains the joint claim-abstract representations and perform rationale selection with the Longformer model \citep{Beltagy2020Longformer}, a transformer model for long documents that takes up to 4096 tokens.

\subsection{Verdict Prediction}
The final step of the fact-checking pipeline is for a model to produce the verdict on a given claim's veracity. As mentioned in the datasets section, the most common setting is to have three labels (\textsc{Supported}, \textsc{Refuted}, \textsc{Not Enough Information}), although models developed for one set of labels can be adapted to a dataset with a different set of labels. This component can easily be modeled as a classification task where the classifier learns to predict one of the three classes. All the baselines from Table \ref{tab:models} perform this task by fine-tuning large language models for label prediction on their respective datasets. The base models used include the general-purpose BERT or T5 and the domain-specific BioBERT and SciBERT \citep{beltagy-etal-2019-scibert} models. These models receive as their input pairs of claims and accompanying rationale sentences selected in the previous step and then give the final output as output.

As described previously, the joint models developed for solving \textsc{SciFact} use multi-task learning to learn both the evidence selection and verdict prediction steps with a shared claim+abstract representation. Both ParagraphJoint and ARSJoint again use a dedicated MLP that takes the previous step's representation. At the same time, ParagraphJoint also experimented with Kernel Graph Attention Network (KGAT), which performed well for general fact-checking datasets by learning relations between evidence sentences using a graph structure \citep{liu-etal-2020-fine}. MultiVerS once again uses the Longformer model, this time with a three-way classification head over the encoding of the entire claim and rationale sentences. 

The MultiVerS model was also used in a zero-shot setting by \citet{wuhrl-klinger-2022-entity} to fact-check the \textsc{CoVERT} dataset. Since this dataset consists of tweets and is pretty noisy when compared to expert-written claims found in \textsc{\textsc{SciFact} }, the authors transformed the tweets into atomic claims consisting of triples (entity, cause, entity). Such a representation significantly improved the performance on this dataset and showed that models developed for one scientific fact-checking dataset can provide promising results for other datasets when the claims are represented in an appropriate form.

\section{Discussion}

In this chapter, we discuss the current challenges in scientific fact-checking and provide directions for future work and trends.

\textbf{Evidence quality.} A common challenge in fact-checking is ensuring that the evidence used for making veracity decisions is appropriate and of high quality. Especially in scientific fact-checking, the nature of scientific knowledge is such that it is updated and readjusted as new discoveries appear, so a claim that was once refuted by evidence could become supported with more substantial, more recent evidence. Time-aware scoring for evidence ranking was explored for general fact-checking \cite{ALLEIN2021100663}. Additionally, scientific sources can contradict one another and give differing results for the same research hypotheses, which is related to the ML concept of learning with label disagreement \citep{uma2021learning}. In the medical field, systematic reviews provide evidence-based clinical recommendations with the level of evidence (how much testing was performed) and the strength of recommendation (is it just a hint or a strict medical recommendation) \citep{cro2020sensitivity}. So far, none of the datasets have taken into account the evidence that is changing with time, disagreeing evidence, or differing levels and strength of evidence. A promising research direction is constructing resources and benchmarks that would consider these intricacies of scientific fact-checking. 

\textbf{Reasoning and Explainability.} Fact-checking is one of the NLP tasks where making the models and their decision process transparent and explainable to humans is of high importance for their wide-scale adoption \citep{c7fe925965844640af62141c6f624b96}. Modern deep neural models for NLP tasks are generally described as black-box models, and their inner workings are still hard to grasp completely. While there have been explainable approaches for general fact-checking, the only explainable method in this survey was proposed by \citet{kotonya-toni-2020-explainable}. It uses a combination of extractive and abstractive text summarization of evidence source documents to provide end users with a concise explanation of why a certain verdict was produced. Considering that scientists often present their thoughts with argumentative structures \citep{lauscher-etal-2018-argument}, a promising research approach is learning the conceptual relations between multiple pieces of evidence to come up with a conclusion. This was used by \citet{krishna2022proofver} to develop a neuro-symbolic model that learns logical relations between evidence sentences for \textsc{Fever}. Another promising research avenue is using counterfactual explanations, which have proven useful in many NLP tasks \citep{ijcai2021p609}.

\textbf{Dataset size.} A common obstacle in fact-checking for all domains and related misinformation detection tasks is the small size of existing datasets. One way to overcome this performance hindrance is combining multiple scientific fact-checking datasets or datasets for related NLP tasks that deal with seeking rationale in text. The model MultiVerS described in the previous chapter utilized this approach by combining datasets \textsc{HealthVer}, \textsc{COVID-Fact}, and \textsc{SciFact} together with FEVER, PubMedQA, and EvidenceInference datasets to improve the final performance on the fact-checking task. Other than combining datasets for training purposes, another emerging approach to mitigate the lack of training data is generating new scientific claims to augment the existing data. \citet{wright-etal-2022-generating} apply this approach by using the generative model BART 
and external biomedical knowledge sources to construct claims while showing promising zero-shot performance.

\textbf{External knowledge.} Scientific knowledge is complex and contains lots of interconnected concepts. This makes it suitable for representation with structures like Knowledge Graphs (KGs) that model world knowledge in the form of entities and relations between them. KGs have been constructed for various scientific disciplines, while the most well-known one for biomedical knowledge is Unified Medical Language System (UMLS) \citep{umls}, which models various interactions between proteins, drugs, diseases, genes, and other concepts. KGs have proven useful in enhancing a wide array of NLP tasks \cite{schneider-etal-2022-decade}. Enhancing BERT with infused disease knowledge from MeSH \citep{he-etal-2020-infusing} and structured medical knowledge from UMLS \citep{he2020bert} showed improved performance over knowledge-intensive biomedical NLP tasks, as well as for the open-domain question answering \citep{yu2021kg}. Recent work has shown that reasoning over knowledge graphs can improve encyclopedic fact verification \cite{kim2023factkg}.


\textbf{Multimodality and multilinguality.} Misinformation is increasingly being spread in forms other than text, including misleading images, artificially constructed videos, or incorrect figures \citep{10.1145/3477495.3531744}. Visuals were an especially popular tool for spreading misinformation about the COVID-19 pandemic \citep{brennen}. Particularly in scientific publications, authors present their data in the forms of figures, tables, and other visualizations. The FEVEROUS shared task \citep{aly-etal-2021-fact} made progress in this direction by requiring participants to develop systems that verify claims over evidence in the structured format (tables and lists). Other than multiple modalities, online claims are made in a multitude of world languages, which calls for the development of efficient multilingual models for scientific fact-checking.

\textbf{Human-centered fact-checking.} Most of the developed fact-checking systems are still limited in practical use because their system design often does not take into account how fact-checking is done in the real world \cite{glockner-etal-2022-missing} and ignores the insights and needs of various stakeholder groups core to the fact-checking process \cite{10.1145/3555143}. Several works started to investigate human evaluation in fact-checking systems. Examples include effectively delivering the misinformation detection results to users \cite{10.1145/3292522.3326012} or guiding the user toward fact-checked news \cite{10.1145/3485447.3512246}. Making the process of NLP-based fact-checking more human-centered is a promising future direction that will make it more reliable, trustworthy, and easier for wide-scale adoption.

\section{Conclusion}
In this survey, we reviewed and systematized existing datasets and solutions for the task of scientific fact-checking. We introduced the task and compared it to its related NLP endeavors, described the existing datasets and their construction process, and explained the models used for scientific fact-checking with their pipeline components. Finally, we provided a critical discussion of current challenges and highlighted promising future directions for the task of scientific fact-checking.

\section{Limitations}
Even though we performed a rigorous literature search to try to cover all existing work on scientific fact-checking, there is possibly work that was left uncovered due to different keywords, naming conventions (e.g., fact-checking vs. claim verification). Whenever possible, we tried covering all related work and all relevant cited papers.

All approaches for automated scientific fact-checking described in this work are still not safe for widespread adoption in practice due to constraints to their performance. Having deployed automated fact-checking systems that would produce incorrect verdicts could lead to mistrust in their usefulness and the process of fact-checking itself, including the work of dedicated manual fact-checkers.

\section*{Acknowledgements}
This research has been supported by the German Federal Ministry of Education and Research (BMBF) grant 01IS17049 Software Campus 2.0 (TU München). We would like to thank the anonymous reviewers for their helpful feedback.

\bibliography{anthology,custom}
\bibliographystyle{acl_natbib}

\end{document}